\documentclass{article}
\usepackage{ijcai24}

\usepackage{times}
\usepackage{soul}
\usepackage{url}
\usepackage[hidelinks]{hyperref}
\usepackage[utf8]{inputenc}
\usepackage[small]{caption}
\usepackage{graphicx}
\usepackage{amsmath}
\usepackage{amsthm}
\usepackage{booktabs}
\usepackage{algorithm}
\usepackage{algorithmic}
\usepackage[switch]{lineno}

\usepackage{algorithm}
\usepackage{algorithmic}
\usepackage{graphicx}
\usepackage{amsmath}
\usepackage{amssymb}
\usepackage{booktabs}
\usepackage{subcaption}
\usepackage{multirow}
\usepackage{stfloats}

\title{Enhancing the \textquotedblleft Immunity\textquotedblright \  of Mixture-of-Experts Networks for Adversarial Defense}

\author{
Qiao Han$^1$
\and
yong huang$^1$\and
xinling Guo$^1$\and
Yiteng Zhai$^1$\and
Yu Qin$^2$\And
Yao Yang$^{1*}$
\\
\affiliations
$^1$Zhejiang Lab\\
$^2$Institute of Software, Chinese Academy of Science\\
\emails
\{hanq, yhuang, guoxl, ito, yangyao\}@zhejianglab.com,
qinyu@iscas.ac.cn
}
\begin{document}

\maketitle

\begin{abstract}
Recent studies have revealed the vulnerability of Deep Neural Networks (DNNs) to adversarial examples, which can easily fool DNNs into making incorrect predictions. To mitigate this deficiency, we propose a novel adversarial defense method called \textquotedblleft Immunity\textquotedblright \  (Innovative MoE with MUtual infomation \& positioN stabilITY) based on a modified Mixture-of-Experts (MoE) architecture in this work. The key enhancements to the standard MoE are two-fold: 1) integrating of Random Switch Gates (RSGs) to obtain diverse network structures via random permutation of RSG parameters at evaluation time, despite of RSGs being determined after one-time training; 2) devising innovative Mutual Information (MI)-based and Position Stability-based loss functions by capitalizing on Grad-CAM's explanatory power to increase the diversity and the causality of expert networks. Notably, our MI-based loss operates directly on the heatmaps, thereby inducing subtler negative impacts on the classification performance when compared to other losses of the same type, theoretically. Extensive evaluation validates the efficacy of the proposed approach in improving adversarial robustness against a wide range of attacks.
\end{abstract}

\section{Introduction}

Deep neural networks (DNNs) have demonstrated remarkable capabilities and ubiquitous applications across various domains, including medical diagnosis \cite{kleppe2021designing}, autonomous driving \cite{stocco2022thirdeye}, security surveillance \cite{amrutha2020deep}, etc. However, recent studies have revealed the vulnerability of DNNs to adversarial attacks \cite{xie2020adversarial}. Adversarial attacks craft imperceptible perturbations into the original input data and induce DNNs to generate erroneous outputs or predictions  \cite{chilamkurthy2018deep}. These attacks not only undermine the effectiveness and reliability of DNNs, but also pose risks to safety-critical scenarios involving these models, such as medical misdiagnosis, traffic accidents, and information leakage \cite{papernot2017practical}. Therefore, enhancing the adversarial robustness of DNNs, i.e. strengthening these models against diverse adversarial attacks, is an important yet challenging problem.

To date, a myriad of adversarial defense strategies have emerged to address this problem, which can be principally classified into 3 main approaches: robust optimization, data compression and auxiliary means \cite{yuan2019adversarial}. Robust optimization trains the model using adversarial examples or regularizers to enhance adaptability and generalizability \cite{madry2017towards} \cite{tramer2017ensemble}. Data compression alleviates harmful data and reduces perturbations \cite{guo2018lau}. Auxiliary means incorporate additional mechanisms to improve model complexity and reliability \cite{margeloiu2020improving}. However, extant methods exhibit certain limitations. For instance, robust optimization risks overfitting or compromising accuracy \cite{kurakin2016adversarial}, data compression may forfeit useful information or features \cite{zhang2019theoretically}, while auxiliary means can impose computational overhead or have limited applicability \cite{dimitrov2020provably}. Moreover, existing methods do not fully consider how model architecture and behavior influence robustness, or how different input data components contribute to robustness.

To address the limitations of prior methods, we propose a novel approach termed \textquotedblleft Immunity\textquotedblright \ (Innovative MoE with MUtual info \& positioN stabilITY) that enhances model defense against adversarial attacks.  Immunity enhances model robustness by employing multiple sub-models termed expert networks to disentangle distinct features and extract unique concepts or semantics. The key innovation in our Mixture-of-Experts (MoE) architecture is integrating a Random Switch Gate, instead of a deterministic gate, to coordinate and aggregate outputs from individual experts. During inference, it randomly exchanges weights to improve robustness. To further enhance multi-angle learning, we devise an innovative Mutual Information (MI) loss and position-stability loss based on Grad-CAM. These losses directly regularize heatmap of expert network to increase diversity and causality of learned representations.

The core innovations and contributions of this paper are:
\begin{itemize}
    \item We propose a novel \textquotedblleft Immunity\textquotedblright \ approach that incorporates Random Switch Gates into MoE architecture to jointly enhance adversarial robustness and interpretability, providing new insights on leveraging model ensemble diversity for defense.
    \item We devise an innovative MI-based loss operating on Grad-CAM, which simplifies calculation and induces subtler negative impact on predictions compared to prevailing MI losses. We also formulate a position-stability regularizer that encourages learning of causal representations by expert networks.
    \item Through extensive experiments on public datasets, we demonstrate  Immunity significantly outperforms state-of-the-art defense methods against diverse adversarial attacks under both standard and adversarial training settings. Our approach exhibits consistent superiority in defending against various attack types and magnitudes.
\end{itemize}

The remainder of this paper is organized as follows: Section \ref{section:rw} reviews related prior works. Section \ref{section:bm} provides background and motivations. Further, Section \ref{section:me} elaborates on the principles and details of the proposed method “Immunity”. In Section \ref{section:ex}, we present experimental setup and results in detail. Finally, Section \ref{section:co} concludes the paper.

\section{Related Works} \label{section:rw}
\subsection{Adversarial Attacks and Defences}
Experiments have revealed deep neural networks' intrinsic vulnerability to adversarial examples, which can easily lead to misclassifications 
 \cite{Goodfellow2014ExplainingAH}. Since the seminal work in 2014, numerous attack techniques have emerged, including the pioneering Fast Gradient Sign Method (FGSM) that manipulates gradient directions to generate attacks 
 \cite{Goodfellow2014ExplainingAH}. Subsequent methods further leverage Jacobian matrices, decision boundaries, and generative adversarial networks. Notable attack algorithms include the Basic Iterative Method(BIM) \cite{Kurakin2016AdversarialEI}, Momentum Iterative Method(MIM) \cite{Dong2017BoostingAA}, C\&W attack \cite{Carlini2018AudioAE} and Projected Gradient Descent (PGD) attack \cite{Madry2017TowardsDL}.

In response, defense strategies have been explored from 3 primary perspectives to enhance model robustness and alleviate vulnerability against adversarial attacks: Robust Optimization, Data Compression, and Auxiliary Technique. Our major focus is on robust optimization
% , including regularization, adversarial training, and certified defenses
. For instance, Kannan et al. \cite{kannan2018adversarial} incorporated adversarial examples into model training, making it the most ubiquitous and fundamental defense approach.Concurrently, Liu et al. \cite{liu2018towards} introduced random modules and proposed Robust Neural Networks via Random Self-Ensemble (RSE) to defend against attacks. Then Hierarchical Random Switching\cite{wang2019protecting} was leveraged for Protecting Neural Networks. 
% Other representative methods include integrating neural architecture search for adversarial robust architectures by \cite{dong2019neural}. 
Consequently, to further improve the search algorithm framework, Robnets \cite{guo2019meets} and AdvRush \cite{Mok2021AdvRushSF} have been proposed to achieve superior robustness. Regarding the data perspective, Jia et al. proposed ComDefend using image compression to eliminate perturbations \cite{jia2019comdefend}, while Prakash et al. generated samples via local corruption \cite{prakash2018deflecting}. Auxiliary techniques include adversarial detection \cite{abusnaina2021adversarial} and auxiliary blocks \cite{yu2019auxblocks}.

\subsection{CAM-based Explanations for Deep Learning Interpretability}
CAM-based methods provide visual explanations for deep learning models. This started with CAM \cite{zhou2016learning}, which required a global pooling layer. Grad-CAM \cite{selvaraju2017grad} then generalized CAM to all neural networks by using gradient information. As an alternative approach, Score-CAM \cite{wang2020score} and SS-CAM \cite{wang2020ss} weighted activation maps based on forward pass scores. In our proposed approach, we utilize Grad-CAM to visualize the internal mechanisms of the model, and leverage this to constrain the learning patterns of individual expert networks.

\subsection{Mutual Information Aided Mixture of Experts for Independent Representations}

Mutual information (MI) and Mixture of Experts (MoE) frameworks have emerged as pivotal techniques for learning independent, disentangled representations across multiple dimensions. Initial explorations into MI date back to Bengio's \cite{bengio2017consciousness}  proposed architecture The Consciousness Prior, which sought to leverage MI for representation learning. Subsequently, Brakel and Bengio formally extracted statistically independent components by minimizing MI \cite{brakel2017learning}. MI has also been utilized for learning disentangled representations \cite{sanchez2020learning}. Recently, Zhou et al. enhanced adversarial robustness by comparing MI between model outputs and natural patterns \cite{zhou2022improving}. 
However, accurately estimating MI remains challenging for high dimensional variables. To overcome this limitation, Belghazi et al. proposed estimating MI based on the Donsker-Varadhan representation of the Kullback-Leibler divergence \cite{belghazi2018mutual}, while Hjelm et al. introduced a Jensen-Shannon divergence objective for Deep InfoMax models \cite{bachman2019learning}.
The Mixture of Experts framework \cite{masoudnia2014mixture}, since its proposal, has achieved remarkable success. Recent works have focused on modeling expert relationships, including Top-k gating \cite{shazeer2017outrageously},  heuristic combinations \cite{wang2018skipnet} and DSelect-k \cite{hazimeh2021dselect}. 
% MI and MoE are both important components of multi-dimensional independent representation.  As early as 2017, bengio\cite{bengio2017consciousness} had similar expressions about mutual information when he proposed the research framework in \emph{The Consciousness Prior}. Then Philemon Brakel\cite{brakel2017learning} proposed to extracted statistically independent components from data based on a minimization of the mutual information. Further it was used for learning disentangled representations\cite{sanchez2020learning}. Zhou\cite{zhou2022improving} compared MI between outputs and natural pattern and MI between outputs and natural pattern to improve adversarial robustness. However, the mutual information is notoriously hard to compute for high-dimensional variables in practice. To solve it, Belghazi\cite{belghazi2018mutual} estimated mutual information based on the Donsker-Varadhan representation of the Kullback-Leibler divergence. Similarly, Hjelm\cite{bachman2019learning} propose a Jensen-Shannon divergence objective function in Deep InfoMax model.  
% Mixture-of-Experts (MoE)\cite{masoudnia2014mixture} structure has achieved great success since it was proposed. Recent work is about expert relationships like Top-k gate\cite{shazeer2017outrageously}, 
% DSelect-k\cite{hazimeh2021dselect} and heuristics\cite{wang2018skipnet}. 
% Our approach combines MI and MoE structure for multi-angle learning to improve the adversarial robustness.
Our method combines MI and MoE structures in an organic way to improve the adversarial robustness: the MoE architecture provides the possibility of multi-angle learning, while MI, as one of the auxiliary loss functions, further diversifies the learning expression of each sub-expert network.

\begin{figure*}[t]
\centering
\includegraphics[width=1.0\textwidth]{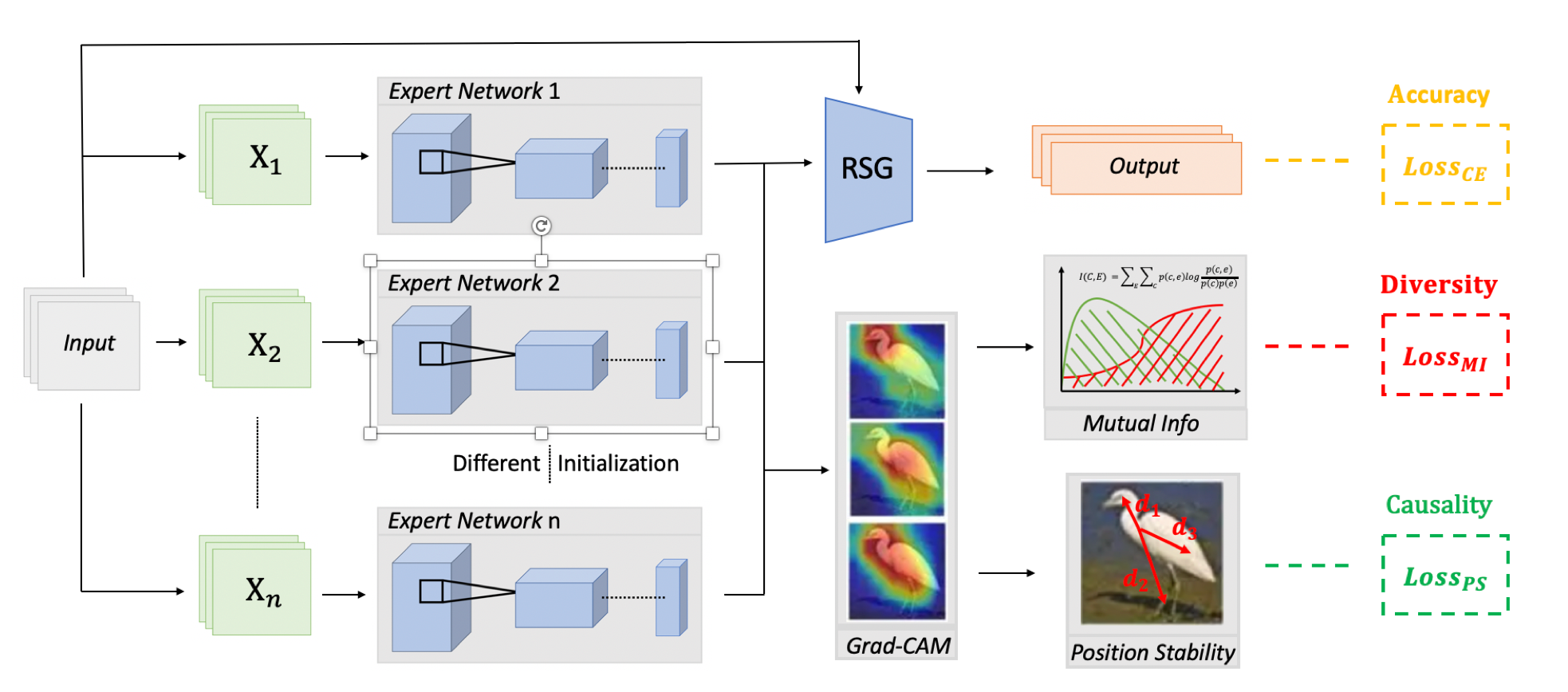} % Reduce the figure size so that it is slightly narrower than the column.
\caption{The Architecture of The Proposed \textquotedblleft Immunity\textquotedblright}
\label{fig:architecture}
\end{figure*}

\section{Background and Motivation} \label{section:bm}
Computer vision has been profoundly impacted by adversarial attacks, because images contain abundant redundant information which can provide attackers with attack materials to fool models into incorrect predictions. In this paper, we primarily focus on the issue of adversarial defense in image classification tasks, and demonstrate how our proposed model and other state-of-the-art defenses perform against various adversarial attacks. Take the most common attack method, FGSM and PGD as examples to describe the principle of attack algorithms:
% Computer vision has been one of the most affected fields in artificial intelligence by adversarial attacks. However, a random noise will hardly lead the model to incorrect predicted results. Most effective adversarial attacks must involve adding deliberate perturbations to the original input data in the anti-gradient direction. Take the most common attack method, FGSM and PGD as examples:

\emph{Fast Gradient Sign Method} (FGSM). An efficient attack based on the gradient of the cost function w.r.t. the neural network input.  $X^*=X-\alpha*sign(\nabla_xloss(x,y^{label})$.

\emph{Projected Gradient Descent} (PGD). A more powerful extension of FGSM via multi-step variant, $X_0 = X, X^{t+1}=\prod_{X+S}{(X^t-\alpha*sign(\nabla_{x^t}loss(x^t,y^{label})))}$, where $\prod_{X+S}$ indicates projection to an allowed perturbation set $S$, which is a ball centered at $X$.

According to the above principle, then we can explain our key motivation why an MoE-based multi-angle learning approach with a Random Switch Gate can provide effective defense. This answer is based on two premises: (i) Most effective adversarial attacks must involve deliberately perturbing the original input in an anti-gradient direction while random noise is unlikely to mislead models. (ii) A single image contains a vast amount of redundant information for classification tasks. Consequently, this redundancy leads to significant divergence in learned parameters depending solely on model initialization. Studies demonstrate models with identical architectures exhibit distinct vulnerable spots when trained from different initializations. Given adversarial attacks are linked to a special vulnerable direction, and model vulnerability varies with random initialization, successfully attacking the MoE architecture requires overcoming the randomization rised by multi-angle learning to identify a universal weakness. This is inherently more challenging. Our mutual information and position stability losses further increase difficulty by promoting expert diversity. 
Additionally, the interpretability afforded by the MoE structure plausibly improves robustness, as interpretability is often positively associated with security.
% Our paper will demonstrate how our model and other state-of-the-art defense models perform against these and other adversarial attacks. To revisit our motivation, why a MoE-based multi-angle learning model with a Random Switch Gate can be an effective defense against these attacks? As is well known, a single image contains a vast amount of redundant information to derive classification results. This leads to significant differences in the learned parameters of models with different initialization. Recent studies have also shown that models with the same network architecture will have the different vulnerable spots when trained with the different initial weights. Since adversarial attacks are linked to worst-case performance, and model vulnerability varies with initial randomization, a successful attack on a MoE-based multi-angle learning model with a Random Switch Gate requires finding a common weakness that applies to all sub-model variants. It is strictly more difficult and our mutual-information-based loss and position-stability-based loss further increases the difficulty. We can also explore this issue from another perspective -using a MoE-based structure makes the model more interpretable and transparent. The interpretability is often positively correlated with the model's security.

\section{Methodology} \label{section:me}

% This section describes the core components of our model to improve the adversarial robustness of neural networks. Our work is relies on a modified MoE structure, along with two constraints: mutual information and position stability index. We will introduce them in detail accordingly. Figure \ref{fig:architecture} is the overall model architecture.
This section describes the core components of our proposed model for improving adversarial robustness of neural networks. Our approach relies on a modified Mixture-of-Experts architecture, along with 2 key constraints: mutual information and position stability regularization. Each of these techniques is described in detail. Figure \ref{fig:architecture} illustrates the overall architecture of our model.

\subsection{Enhancing Mixture of Experts Robustness via Novel Regularizations}

A standard Mixture of Experts (MoE) learns from a set of expert networks $f_i$ through a gating network $g$, where $i = 1,2......N$ and $N$ is the number of expert networks. The expert networks comprise standard CNNs like ResNets and GoogLeNets. Although each expert network takes the same input $x$, we refer to the inputs as $x_i$ for notation convenience, where each xi is a duplicate of $x$. The final output is:
\begin{equation} 
\label{eq_m1} 
\begin{split}
&\hat{y} = \sum^{N}_{i=1}{g_i(x)\hat{y}_i} 
\\&where  \ \ \hat{y}_i=\mathrm{softmax}(f_i(x_i))
\end{split}
\end{equation}
Thus we obtain the first loss function based on cross-entropy to optimize the classification accuracy.
\begin{equation} 
\label{loss1} 
\resizebox{0.95\linewidth}{!}{$
        \displaystyle
        loss_{CE} = -\sum^{M}_{m=1}{y^{label}_mlog(\hat{y}_m)} - \alpha\sum^{N}_{i=1}{\sum^{M}_{m=1}{y^{label}_mlog(\hat{y}_{im})}}
        $}
\end{equation}
% where $m$ indexes categories. Through the aforementioned loss function, we are essentially conducting supervised training for each expert network with the same label. The gate can be considered as the weight of each expert network's outputs. Although it is determined during once training and used to generate adversarial attack samples, we can obtain various architectures during inference process by randomly switch the weights. Since each expert network has received sufficient training, introducing the Random Switch Gate will not have a negative impact on the classification results. It can effectively defend against adversarial attacks by preventing attackers from attacking each expert with fixed weights. At the same time, the structure of RSG is simple enough to prevent the problem of unstable prediction results encountered in complex Neural Architecture Search algorithms.

where $m$ indexes categories. This loss trains each expert network in a supervised manner using the same labels. Further, the gate weights the expert network’s outputs. Although determined after one-time training, we randomly switch the gate parameters during inference to obtain diverse architectures. As expert networks receive sufficient training, introducing the Random Switch Gate does not impact classification but prevents attackers from targeting fixed weights. Moreover, the simple structure of RSG also avoids instability encountered in complex Neural Architecture Search algorithms.

In MoE, each expert learns different higher-level representation. However, without external constraints, the distinctions between expert representations are not readily apparent or pronounced. To enhance model robustness, we impose constraints to achieve:
(i) Unique expert learning basis - for example, in bird classification, experts may focus on different components like wings, claws and feathers. Once these judgments are combined and supervised by the gating network, stable outputs can be maintained even if one component is attacked.
(ii) Causal representations - for instance, high-flying birds correlate with clouds but may not imply causation. Adversarial attacks often exploit such non-causal spurious correlations between input patterns and model outputs. By constraining experts to learn more robust causal representations and thus enhance model security, we aim to remove potentially vulnerable non-causal links. 
We consequently design the MoE framework based on these principles to enhance adversarial robustness.

\subsection{Interpreting Experts through Grad-CAM }

% In order to enhance the diversity and causality of learning basis, we firstly use the Grad-CAM method to determine the importance of each pixel for an expert's decision. In our model, the Grad-CAM method is applied on the last convolutional layer $A_i$ whose size is larger than 8*8. We conducted several trials and expect to have the best compromise between high-level semantics and detailed spatial information. For the class given by label, Grad-CAM averages the gradients of the output $y^{label}$ with respect to all channels to indicate its importance. The heatmap of $A_i$ is
In order to promote diversity and causality of learned representation in experts, we first leverage Grad-CAM to determine each pixel's importance for an expert's output decision. In our model, Grad-CAM is applied to the last convolutional layer $A_i$ with dimensions exceeding $8*8$, chosen after experimentation to balance high-level semantics and spatial details. For a given class label, Grad-CAM averages the gradient of the label output $y^{label}$ w.r.t. all channels, indicating importance. The heatmap of $A_i$ is

\begin{equation} 
\label{eq_m2} 
\begin{split}
&H^{label}_i = ReLU(\sum^{M}_{k=1}{\alpha^k_iA^k_i})
\\& with \ \ \  \alpha^k_i = \frac{1}{M} \sum_{u,v} \frac{\partial y^{label}}{\partial A^k_{i(u,v)}}
\end{split}
\end{equation}
 where $k = 1,2......M$ indexes channels and $(u,v)$ represents the denotes feature coordinates in $A_i$. Then we reduce $h_i$, the heatmap of $x_i$, by linear interpolation of $H^{label}_i$, the activation heatmap of $A_i$. Despite all experts receiving virtually identical inputs $x_i$ , we aim to generate unique heatmaps $h_i$  for each individual expert network.

 \subsection{Simplified MI for Expert Diversity}
% To improve the diversity of experts' learning basis, the most straightforward solution is to minimize the mutual information(MI) between the learned high-level representations of individual expert networks. However, this idea has three drawbacks: (i) Calculating exact global MI of learned representations is complicated and challenging. (ii) High-level features are intermediate results that implicitly contain information about the final outputs. Thus, constraining the MI between these features might negatively impact output accuracy, unlike our approach. (iii) Fundamentally, this method does not align with our starting point of allowing expert networks to learn independently from different original pixel features.

A straightforward approach to improve expert diversity is minimizing the mutual information between learned high-level representations for each individual expert networks. However, this has several limitations: (i) Basically, accurately calculating global MI of learned representations is challenging; (ii) As high-level features are intermediate representations containing implicit info about final outputs, constraining their MI could negatively impact accuracy; (iii) Additionally, minimizing MI between full representations fundamentally misaligns with our goal of enabling experts to learn distinct representations from original pixel-level.

% Consider to randomly extract a pixel weighted by the intensity of its influence on the final output. Its coordinates $C$ whose probability distribution we will denote $p(c)$. We also define a variable $E$ which duplicated inputs the pixel originates. Our approach is to maximize the MI between $C$ and $E$. As MI is a Shannon entropy-based measure to quantify the dependence of two random variables, The higher the MI between $C$ and $E$, the more correlated $E$ and $C$ are, and finally the more unique features each expert network learns from. In mathematics, MI between $C$ and $E$ is defined as
% \begin{equation} 
% \label{eq_m3} 
% \begin{split}
% I(C,E) = \sum_{E}{\sum_{C}{p(c,e)log\frac{p(c,e)}{p(c)p(e)}}}
% \end{split}
% \end{equation}
% Note that $p(c,e)=p(c|e)p(e)$ and $p(c)=\sum_{E}{p(c|e)p(e)}$, then we can obtain

Instead, we propose to maximize $I(C, E)$, the MI between two variables: $C$, a randomly sampled pixel (its coordinate) weighted by output influence; and $E$, the expert network this pixel originates from. 
\begin{equation} 
\label{eq_m3} 
\begin{split}
I(C,E) = \sum_{E}{\sum_{C}{p(c,e)log\frac{p(c,e)}{p(c)p(e)}}}
\end{split}
\end{equation}
MI is a Shannon entropy-based measure to quantify the dependence of two random variables, thus the higher $I(C, E)$, the stronger the correlation between $C$ and $E$, indicating more distinct features learned by each expert network. Note that $p(c,e)=p(c|e)p(e)$ and $p(c)=\sum_{E}{p(c|e)p(e)}$, then we can obtain

\begin{equation} 
\label{eq_m4} 
\begin{split}
I(C,E) &= \sum_{E}{\sum_{C}{p(c,e)log\frac{p(c,e)}{p(c)p(e)}}}
\\&= \sum_{E}{\sum_{C}{p(c|e)p(e)log\frac{p(c|e)}{p(c)}}}
\\&= \frac{1}{N}\sum_{E}{\sum_{C}{p(c|e)log\frac{p(c|e)}{\sum_{E}{(p(c|e)p(e))}}}}
\\&= \frac{1}{N}\sum_{i}{\mathbb{E}_{E=i}log\frac{Np(c|E=i)}{\sum_{j 
}{p(c|E=j)}}}
\end{split}
\end{equation}
% We leverage heatmaps $h_i$ to simulate the distribution $p(c|E=i)$. According to equation (\ref{eq_m4}), Maximizing MI between $C$ and $E$ is roughly equivalent to maximizing the mathematical expectation of the distance from each heatmap $l_i$ to their mean. 

We approximate the distribution $p(c|E=i)$ by using heatmaps $h_i$ as a surrogate. According to Equation (\ref{eq_m4}), maximizing $I(C, E)$ is approximately equivalent to maximizing the mathematical expectation of the distance between each heatmap $h_i$ and their mean.

% We can provide an intuitive interpretation of the theoretical maximum in equation (\ref{eq_m3}) through the following example. Consider to divide $x$ into N subregions and Each subregion has a uniform heatmap ($x$ can be viewed as consisting of n pixels), then equation (\ref{eq_m3}) can be re-written as

An intuitive interpretation of the theoretical maximum in Equation (\ref{eq_m3}) can be provided through the following example. Consider dividing input $x$ into N uniform subregions,and each subregion is assigned a uniform heatmap  ($x$ can be viewed as consisting of N pixels). Under this formulation, Equation (\ref{eq_m3}) can be rewritten as:

\begin{equation} 
\label{eq_m5} 
\begin{split}
I(C,E) = &\frac{1}{N}\sum_{i}{\mathbb{E}_{E=i}log\frac{Np(c|E=i)}{\sum_{j 
}{p(c|E=j)}}}
\\= \frac{1}{N}\sum_{i=2}^N \sum_{k=1}^N & p(c_k|E=i)
\\&log\frac{Np(c_k|E=i)}{p(c_k|E=1)+\sum_{j \neq 1}{p(c_k|E=j)}}
\\+ \frac{1}{N}\sum_{k=1}^N & p(c_k|E=1)
\\&log\frac{Np(c_k|E=1)}{p(c_k|E=1)+\sum_{j \neq 1}{p(c_k|E=j)}}
\end{split}
\end{equation}

Assume $l=\operatorname*{argmin}_k \sum_{j \neq 1}{p(c_k|E=j)}$, we artificially construct a distribution $p^*(c_k|E=1)$ satisfying:

\begin{equation}
\label{eq_m6} 
p^*(c_k|E=1) =\begin{cases}
1 & if \  k=l  
\\0 & if \  k \neq l 
\end{cases}
\end{equation} 

Under this formulation, we can derive an upper bound on Equation (\ref{eq_m5})
\begin{equation} 
\label{eq_m7} 
\begin{split}
&I(C,E) 
% \leq \frac{1}{N}\sum_{i=2}^N({\sum_{k \neq l}{p(c_k|E=i)log\frac{Np(c_k|E=i)}{\sum_{j 
% \neq 1}{p(c_k|E=j)}}}}
% \\& +p(c_l|E=i)log\frac{Np(c_l|E=i)}{1+\sum_{j 
% \neq 1}{p(c_l|E=j)}})
% \\&+ \frac{1}{N}(\sum_{k=1}^N{p(c_k|E=1)})log\frac{Np(c_k|E=1)}{p(c_k|E=1)+\sum_{j \neq 1}{p(c_l|E=j)}}\\& 
\leq \frac{1}{N}\sum_{i=2}^N({\sum_{k \neq l}{p(c_k|E=i)log\frac{Np(c_k|E=i)}{\sum_{j 
\neq 1}{p(c_k|E=j)}}}}
\\& +p(c_l|E=i)log\frac{Np(c_l|E=i)}{1+\sum_{j 
\neq 1}{p(c_l|E=j)}})
\\&+ \frac{1}{N}log\frac{N}{1+\sum_{j \neq 1}{p(c_l|E=j)}}
\\& = \frac{1}{N}(\mathbb{E}_{E=1}log\frac{Np^*(c|E=1)}{p^*(c|E=1)+\sum_{j=2}^N{p(c|E=j)}}
\\& + \sum_{i=2}^N{\mathbb{E}_{E=i}log\frac{Np(c|E=i)}{p^*(c|E=1)+\sum_{j=2}^N{p(c|E=j)}}})
\end{split}
\end{equation}

Equation (\ref{eq_m7}) proves that, when $p(c|E=i) \ \ \forall i\geq2$ is fixed, (\ref{eq_m3}) will reach its maximum if and only if $p(c|E=1)=p^*(c|E=1)$. This implies the $1_{st}$ expert focuses solely on the $l_{th}$ subregion. Similarly, each expert network should learn unique subregions to maximize $I(C, E)$.

Based on the above, we define the second loss function $loss_{MI}$ as
\begin{equation} 
\label{loss2} 
\begin{split}
loss_{MI} = -\sum_{i=1}^N{\sum_{a,b}{h_{i(a,b)}log\frac{h_{i(a,b)}}{\sum_{j=1}^N{h_{j(a,b)}}}}}
\end{split}
\end{equation}
Where $(a,b)$ denotes the coordinates of $h_i$. Obviously, $loss_{MI}$ provides a simplified estimation of the inverse of $I(C,E)$.

\subsection{Regularizing Experts via Position Stability Index}
We incorporate position stability of salient regions as our third loss to promote expert causality. If a pattern consistently triggers an expert's decision of correct result, his position should remain relatively stable across images.

% We incorporate position stability of salient regions as our third loss to promote expert causality.  We assume that if a pattern is a causal credential to derive the correct result, it should be always triggered by the similar positions in different images. Thus, the distances between causal patterns  should remain relatively stable across images.

For each heatmap $h_i$ of expert networks, we computed the \textquotedblleft centre of mass\textquotedblright \ $(X^C_i, Y^C_i)$ as follow
\begin{equation} 
\label{eq_m8} 
\begin{split}
&X^C_i=\frac{\sum_{a,b}{ah_{i(a,b)}}}{\sum_{a,b}{h_{i(a,b)}}}
, \ \ Y^C_i=\frac{\sum_{a,b}{bh_{i(a,b)}}}{\sum_{a,b}{h_{i(a,b)}}}
\end{split}
\end{equation}
Where $(a,b)$ is the coordinate of $h_i$. $(X^C_i, Y^C_i)$ represents the focus of the $i_{th}$ expert network. Based on the above assumptions, if the experts learn causal representations, the distance between their \textquotedblleft centre of mass \textquotedblright \  should not change a lot. Lastly, we define the third loss to minimize the variance of these distance across various images in a mini-batch:
\begin{equation} 
\label{loss3} 
\begin{split}
&loss_{PS} = \sum_{i,j}{\sum_{q=1}^{mb}{{(d_{q\_i,j}^2-\frac{1}{mb}\sum_{q=1}^{mb}{d_{q\_i,j}^2})^2}}}
\\&with \ \  d_{q\_i,j}^2=(X^C_{q\_i} - X^C_{q\_j})^2 + (Y^C_{q\_i} - Y^C_{q\_j})^2
\end{split}
\end{equation}
where $mb$ is the number of sample in a mini-batch.

\subsection{Objective and Training Techniques}
The overall loss function for a mini-batch combines the cross-entropy, mutual information, and position stability losses:
\begin{equation} 
\label{loss} 
\begin{split}
Loss = &\frac{1}{mb}\{\sum_{q=1}^{mb}{loss_{CE}}+\frac{\beta}{N} \sum_{q=1}^{mb}{loss_{MI}}
\\&+\frac{\gamma}{N(N-1)}loss_{PS}\}
\end{split}
\end{equation}
where $loss_{CE}$, $loss_{MI}$, $loss_{PS}$ are defined in equations (\ref{loss1}), (\ref{loss2}), (\ref{loss3}), respectively, and $\beta$, $\gamma$ are multi-objective coefficients. We apply both standard and adversarial training to test their effectiveness separately. In addition, adversarial training is known to be highly effective for defense, and the idea is to use adversarial examples during training and plays a min-max game \cite{kannan2018adversarial}. The inner maximization generates strong adversarial examples that maximize the cross-entropy loss. The outer minimization then updates model parameters to minimize the loss. While adversarial training can significantly improve model accuracy by adversarial samples, it comes at the cost of greatly reduced accuracy on clean examples.
% In addition to standard training, we also apply adversarial training to the model and test their effectiveness separately. Adversarial training\cite{kannan2018adversarial} is known to be the most effective method for defense. This method uses adversarial examples as training data and plays a min-max game. The inner maximum produces stronger adversarial examples to maximize the cross-entropy loss, while the outer minimum updates the model parameters to minimize it. Adversarial training can significantly improve the accuracy of the model on adversarial samples, but at the cost of greatly reducing its accuracy on normal samples.

\section{Experiment} \label{section:ex}

% Our experiments aim to evaluate the robustness and defense potential of the Immunity model against various evasion and poisoning attacks. We aim to answer three main questions:

Our experiments evaluate the robustness and defense capabilities of the proposed Immunity model against various evasion and poisoning attacks. The key questions are: 

\begin{enumerate}
\item Can Immunity exhibit enhanced resilience across diverse attack types? 
\item Which modules - mixture-of-experts, mutual information loss, or position stability loss - most significantly impact Immunity's performance? 
\item How can Grad-CAM's novel visual explanations elucidate Immunity's robustness?
\end{enumerate}

\subsection{Experiment Setting}
\subsubsection{Dataset} We conduct classification on the CIFAR-10 and CIFAR-100 dataset, comprising 32x32 pixel images spanning 10 and 100 classes. Training and test sets are normalized via channel means and standard deviations. We apply standard data augmentation including random crops and rotations.

\subsubsection{Adversarial Attacks} We evaluate both standard and adversarial training. Evasion attacks for robustness quantification include FGSM\cite{Goodfellow2014ExplainingAH}, BIM \cite{Kurakin2016AdversarialEI}, MIM\cite{Dong2017BoostingAA}, PGD\cite{Madry2017TowardsDL} 
Attacks use an 8/255 perturbation scale and 20 iterations for BIM, MIM and PGD attacks, and they are inspired from GitHub\footnote{https://github.com/Harry24k/adversarial-attacks-pytorch}.

\subsubsection{Defence Baselines} Defense baselines are P-DARTS\cite{Chen2019ProgressiveDA}, RobNets\cite{guo2019meets}, AdvRush\cite{Mok2021AdvRushSF} - state-of-the-art methods using regularization. Models are trained per original specifications with default settings in perspective GitHub project page.

 \subsubsection{\textquotedblleft Immunity\textquotedblright} We train Immunity for 200 epochs (batch size 32) under both standard and adversarial regimes, using SGD as optimizer with learning rate 0.01 and weight decay 5e-4. Moreover, we ensemble 5 experts in our experiments, conducted on an Nvidia Tesla V100S PCIE GPU.

 \subsection{Experiment Analysis}

Our experiment employs three prominent deep neural network architectures as the expert models. Subsequently, we select the most proficient architecture as an expert to train the Immunity model. Additionally, we add mutual information and position stability index to calculate the robustness of the models.

Our experiment analysis has four parts. Firstly, we compared the performance of standard CNNs with their corresponding Immunity models to determine the most effective expert for training the Immunity model. Secondly, we perform evaluations to assess the robustness of various models against evasion and poisoning attacks. Thirdly, we explain the robustness of the Immunity network architectures through ablation study. Lastly we visualize our expert Grad-CAM heatmap to explain robustness.

\subsubsection{Key Expert Selection in \textquotedblleft Immunity\textquotedblright} 

We first evaluate three prominent CNNs - ResNet18, DenseNet121, and GoogLeNet on CIFAR-10. These models then serve as expert candidates for training the Immunity model. To fully assess defense capabilities beyond accuracy, we introduce two metrics - Independence Score (IScore) and Causality Score (CScore) based on lossMI and lossPS:

 \textbf{IScore}. Quantifies expert heatmap differences via mean squared intensity deviations, relevant parameters in formula \ref{loss2}:
 
 $$IScore=\frac{1}{N(N-1)}\sum_{i,j}^{N,N}{\sum_{a,b}{(h_{i(a,b)}-h_{j(a,b)})^2}}$$ 

 \textbf{CScore}. Measures expert focus variance across images using center of mass distances, relevant parameters in formula \ref{loss3}:
$$CScore=\frac{1}{mb}\sum_{i,j}{\sum_{q=1}^{mb}{{(d_{q\_i,j}^2 - \frac{1}{mb}\sum_{q=1}^{mb}{d_{q\_i,j}^2})^2}}}$$ 

We simulate experts by training CNNs independently 5 times. Compared to standalone CNNs, Immunity improves IScore and CScore substantially despite a minor accuracy drop, indicating enhanced robustness.For attack defense, we determine the optimal Immunity expert. As shown in Table 1, GoogLeNet achieves the highest accuracy of 95.01\%, surpassing ResNet18 and DenseNet121. Among the Immunity variants, Immunity-GoogLeNet obtains the best accuracy of 93.46\%, superior to Immunity-ResNet18 and Immunity-DenseNet121. Therefore, we select GoogLeNet as the ideal expert for the Immunity model.

\begin{table}[htbp]
  \centering
    \begin{tabular}{l|rrr} 
    \toprule    Model & Accuray & IScore & CScore \\
    \midrule
    ResNet18 & 94.14\% & 0.344 & 0.026 \\
     Immunity-ResNet18 & 91.89\% & 0.125 & 0.003 \\
    \midrule
    GoogLeNet & \textbf{95.01\%} & 0.352 & 0.022 \\
    Immunity-GoogLeNet & \textbf{93.46\%} & \textbf{0.117} & \textbf{0.001} \\
    \midrule
    DenseNet121 & 94.07\% & 0.313 & 0.026 \\
    Immunity-DenseNet121 & 91.73\% & 0.119 & 0.004 \\
    \bottomrule
    \end{tabular}%
    \caption{Evaluation of Candidate Experts for \textquotedblleft Immunity\textquotedblright}
  \label{tab:Selecting-suitable-expert}%
\end{table}%

\subsubsection{Robustness Study: Adversarial Attack Resilience}

We evaluate robustness against prominent FGSM, BIM, MIM, and PGD attacks. The PGD attack is $l_{\infty}$-bounded with total perturbation scale of 8/255 and 20 step iterations. Then we calculate the accuracy of defence models including P-DARTS, RobNets, AdvRush and Immunity model under these various attack modes.

We present attack results on CIFAR-10 and CIFAR-100 under standard training and adversarial training as illustrated in the table ~\ref{tab:my-table-adversarial-attacks}. The conclusion is evident: in the absence of attacks during standard training mode, Immunity exhibits slight lower accuracy compared to part of the baseline models. However, following adversarial attacks, Immunity demonstrates a notable improvement, achieving approximately 5\% to 10\% higher accuracy than other models
% (e.g., Immunity demonstrates an accuracy increase of 10.51\% compared to AdvRush under BIM attack)
. In the adversarial training mode, Immunity significantly outperforms other state-of-the-art methods weather being attacked or not. It demonstrates approximately 2\% to 5\% higher accuracy in a clean test environment and achieves 2\% to 15\% approximately higher accuracy under adversarial attacks
% (e.g., Immunity demonstrates an accuracy increase of 4.89\% compared to RobNets under PGD attack)
, which is notably superior to other baselines.

% \begin{table*}[htbp]
%   \centering
%   \caption{Accuracy of \textquotedblleft Immunity\textquotedblright\ and Baseline Methods Under Attacks with Standard \& Adversarial Training}
%     % Table generated by Excel2LaTeX from sheet 'Sheet4'
%     \begin{tabular}{c|c|ccccc}
%     \toprule
%           & Model &Clean & FGSM & I-FGSM & MI-FGSM & PGD \\
%     \midrule
%     \multirow{3}[2]{*}{Standard  Training} & GoogLeNet & 95.01\% & 47.64\% & 45.39\% & 44.54\% & 38.10\% \\
%           & RobNets & 92.50\% & 50.43\% & 55.76\% & 60.33\% & 46.07\% \\
%           & P-DARTS & 94.61\% & 53.97\% & 55.96\% & 54.88\% & 37.04\% \\
%           & AdvRush & \textbf{97.58\%} & \textbf{60.38\%} & 51.68\% & 56.63\% & 41.55\% \\
%           & \textbf{Immunity} & 93.46\% & 60.24\% & \textbf{62.19\%} & \textbf{61.69\%} & \textbf{54.72\%} \\
%     \midrule
%     \multirow{5}[2]{*}{Adversarial Training} & GoogLeNet & \textbf{89.40\%} & 57.67\% & 56.88\% & 57.23\% & 50.09\% \\
%           & RobNets & 85.66\% & 60.64\% & 59.94\% & 58.86\% & 54.92\% \\
%           & P-DARTS & 83.87\% & 63.50\% & 61.58\% & 41.54\% & 39.91\% \\
%           & AdvRush & 86.38\% & 62.27\% & 58.96\% & 60.84\% & 56.63\% \\
%           & \textbf{Immunity} & 88.25\% & \textbf{64.74\%} & \textbf{64.32\%} & \textbf{62.98\%} & \textbf{59.81\%} \\
%     \bottomrule
%     \end{tabular}%
%   \label{tab:my-table-adversarial-attacks}%
% \end{table*}%

\begin{table*}[htbp]
  \centering
    % Table generated by Excel2LaTeX from sheet 'Sheet4'
    \begin{tabular}{l|l|l|rrrrr}
    \toprule
        Dataset & Training Mode & Model &Clean & FGSM & BIM & MIM & PGD \\
    \midrule
     \multirow{10}[2]{*}{CIFAR-10} & \multirow{5}[2]{*}{Standard  Training} & GoogLeNet & 95.01\% & 47.64\% & 45.39\% & 44.54\% & 38.10\% \\
        &  & RobNets & 92.50\% & 50.43\% & 55.76\% & 60.33\% & 46.07\% \\
        &  & P-DARTS & 94.61\% & 53.97\% & 55.96\% & 54.88\% & 37.04\% \\
        &  & AdvRush & \textbf{97.58\%} & \textbf{60.38\%} & 51.68\% & 56.63\% & 41.55\% \\
        &  & \textbf{Immunity} & 93.46\% & 60.24\% & \textbf{62.19\%} & \textbf{61.69\%} & \textbf{54.72\%} \\
     \cmidrule{2-8}
     & \multirow{5}[2]{*}{Adversarial Training} & GoogLeNet & \textbf{89.40\%} & 57.67\% & 56.88\% & 57.23\% & 50.09\% \\
         & & RobNets & 85.66\% & 60.64\% & 59.94\% & 58.86\% & 54.92\% \\
        &  & P-DARTS & 83.87\% & 63.50\% & 61.58\% & 41.54\% & 39.91\% \\
        &  & AdvRush & 86.38\% & 62.27\% & 58.96\% & 60.84\% & 56.63\% \\
        &  & \textbf{Immunity} & 88.25\% & \textbf{64.74\%} & \textbf{64.32\%} & \textbf{62.98\%} & \textbf{59.81\%} \\
    \midrule
     \multirow{10}[2]{*}{CIFAR-100} & \multirow{5}[2]{*}{Standard  Training} & GoogLeNet & \textbf{76.10\%} & 36.26\% & 31.89\% & 30.48\% & 24.80\% \\
        &  & RobNets & 71.79\% & 46.38\% & 40.03\% & 40.06\% & 33.78\% \\
        &  & P-DARTS & 68.47\% & 48.55\% & 43.11\% & 39.96\% & 34.33\% \\
        &  & AdvRush & 66.53\% & 49.72\% & \textbf{45.15\%} & 41.29\% & 37.15\% \\
        &  & \textbf{Immunity} & 74.01\% & \textbf{51.64\%} & 44.34\% & \textbf{42.09\%}& \textbf{40.62\%} \\
    \cmidrule{2-8}
     & \multirow{5}[2]{*}{Adversarial Training} & GoogLeNet & \textbf{70.93\%} & 47.05\% & 44.03\% & 42.47\% & 38.58\% \\
        &  & RobNets & 67.44\% & 50.26\% & 46.13\% & 44.86\% & 41.89\% \\
        &  & P-DARTS & 66.14\% & 49.17\% & 47.71\% & 45.32\% & 43.11\% \\
        &  & AdvRush & 64.43\% & 49.39\% & 45.15\% & 46.31\% & 44.45\% \\
        &  & \textbf{Immunity} & 68.19\% & \textbf{54.93\%} & \textbf{48.48\%} & \textbf{47.17\%} & \textbf{45.24\%} \\
    \bottomrule
    \end{tabular}%
    \caption{Accuracy of \textquotedblleft Immunity\textquotedblright\ and Baseline Methods Under Attacks with Standard \& Adversarial Training}
  \label{tab:my-table-adversarial-attacks}%
\end{table*}%

\subsubsection{Ablation Study: Evasion Attack Evaluation}

To determine the most critical Immunity component, we evaluate Mixture-of-Experts (MoE) variants on CIFAR-10 under standard training:

\begin{enumerate}
\item MoE: Vanilla model
\item MoE+MI: model with mutual information loss
\item MoE+PS: model with position stability loss
\end{enumerate}

\begin{table}[htbp]
  \centering
    \begin{tabular}{l|rrrr}
    \toprule
          & Clean & FGSM  & BIM & PGD \\
    \midrule
    GoogLeNet & 95.01\% & 47.64\% & 45.39\% & 38.10\% \\
    \midrule
    MoE  & \textbf{94.74\%} & 53.26\% & 55.54\% & 48.44\% \\
    MoE+MI & 93.19\% & \textbf{58.91\%} & \textbf{60.90\%} & \textbf{52.02\%} \\
    MoE+PS & 93.58\% & 55.42\% & 58.99\% & 49.87\% \\
    \midrule
    \textbf{Immunity} & 93.46\% & \textbf{60.24\%} & \textbf{62.19\%} & \textbf{54.72\%} \\
    \bottomrule
    \end{tabular}%
    \caption{Ablation Study for \textquotedblleft Immunity\textquotedblright }
  \label{tab:Ablation-study}%
\end{table}%

As shown in Table 3, under clean training vanilla MoE achieves the highest accuracy of 94.74\%, surpassing MoE+PS and MoE+MI. However, against evasion attacks, MoE+MI exhibits approximately 2-5\% higher accuracy over MoE+PS and MoE. This suggests mutual information loss significantly impacts Immunity's performance, while position stability loss still improves MoE robustness. Both modules are crucial for Immunity's overall robustness gains. But the ablation study highlights mutual information as the most vital component for adversarial defense.

\subsubsection{Visualization Study: Immunity Heatmaps}

We want to visualize Immunity model to explain why our model is robust, hence we generate images and their corresponding expert Grad-CAM heatmaps to highlight the distinctive characteristics of each expert. We put 4 sampled pictures and their Grad-CAM heatmaps into a 4*6 matrix, the first column is a 32*32*3 raw picture in each row and follows five normalized 32*32*3 Immunity expert Grad-CAM heatmaps after standard training.

\begin{figure}[t]
\centering
\includegraphics[]{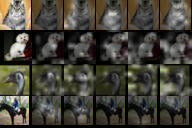} % Reduce the figure size so that it is slightly narrower than the column.
\caption{Visualization of \textquotedblleft Immunity\textquotedblright\  Expert Heatmaps and Original Input Images }
\label{fig:explanation-of-Immunity}
\end{figure}

The intensity of the heatmap corresponds to the importance of each part, with lighter regions indicating higher significance compared to other areas. Looking at the second row in the figure \ref{fig:explanation-of-Immunity}, we can clearly observe that each expert has some different concentrates on exampled picture. Moving from left to right, the expert Grad-CAM heatmaps specialize in capturing details related to the body, ears and hind legs, head, back, tail and front legs of the dog.

\section{Conclusion and Future Works} \label{section:co}

In summary, we have proposed \textquotedblleft Immunity\textquotedblright, a novel regularized Mixture-of-Experts framework strengthened by mutual information and position stability techniques to enhance adversarial robustness. Through extensive experiments on benchmark datasets, we have demonstrated that Immunity significantly outperforms state-of-the-art defenses against a wide array of evasion and poisoning attacks under both standard and adversarial training protocols. The consistent superiority of Immunity highlights the effectiveness of diversifying and regularizing experts within ensemble architectures to withstand adversarial manipulations. This work opens exciting avenues for future research into robust and interpretable multi-gate MoE algorithms that leverage mutual information principles for broader applications involving multi-modal, graph, and other complex data. By combining diversity, causality, and interpretability within ensemble models, this research direction could help develop more robust defenses beyond current security-robustness trade-offs, enabling steady progress toward delivering trustworthy and reliable AI systems.

% \begin{algorithm}[tb]
%     \caption{Example algorithm}
%     \label{alg:algorithm}
%     \textbf{Input}: Your algorithm's input\\
%     \textbf{Parameter}: Optional list of parameters\\
%     \textbf{Output}: Your algorithm's output
%     \begin{algorithmic}[1] %[1] enables line numbers
%         \STATE Let $t=0$.
%         \WHILE{condition}
%         \STATE Do some action.
%         \IF {conditional}
%         \STATE Perform task A.
%         \ELSE
%         \STATE Perform task B.
%         \ENDIF
%         \ENDWHILE
%         \STATE \textbf{return} solution
%     \end{algorithmic}
% \end{algorithm}

%% The file named.bst is a bibliography style file for BibTeX 0.99c
\bibliographystyle{named}
\bibliography{ijcai24}

\end{document}